\documentclass[10pt,twocolumn,letterpaper]{article}

\usepackage{cvpr}
\usepackage{times}
\usepackage{epsfig}
\usepackage{graphicx}
\usepackage{amsmath}
\usepackage{amssymb}
\usepackage[flushleft]{threeparttable}
\usepackage{booktabs}

\usepackage{multirow}
\usepackage{makecell}
\usepackage{tabularx}
\usepackage{xcolor}
\usepackage{footnote}
\usepackage{caption}
\usepackage{textcomp, gensymb}

\usepackage{algorithm}
\usepackage{algpseudocode}

\newcolumntype{M}{>{$}c<{$}}
\newcolumntype{Z}{>{\centering\arraybackslash}X}
\newcolumntype{C}{>{\centering\arraybackslash}p}
\newcolumntype{Y}{>{\centering\arraybackslash}X}
\newcolumntype{L}{>{\raggedright\arraybackslash}p}

\begin{document}

\title{GenPalm: Contactless Palmprint Generation with Diffusion Models}

\author{Steven A. Grosz\\
Michigan State University\\
{\tt\small groszste@msu.edu}
\and
Anil K. Jain\\
Michigan State University\\
{\tt\small jain@msu.edu}
}

\maketitle
\thispagestyle{empty}

\begin{abstract}
   The scarcity of large-scale palmprint databases poses a significant bottleneck to advancements in contactless palmprint recognition. To address this, researchers have turned to synthetic data generation. While Generative Adversarial Networks (GANs) have been widely used, they suffer from instability and mode collapse. Recently, diffusion probabilistic models have emerged as a promising alternative, offering stable training and better distribution coverage. This paper introduces a novel palmprint generation method using diffusion probabilistic models, develops an end-to-end framework for synthesizing multiple palm identities, and validates the realism and utility of the generated palmprints. Experimental results demonstrate the effectiveness of our approach in generating palmprint images which enhance contactless palmprint recognition performance across several test databases utilizing challenging cross-database and time-separated evaluation protocols.
\end{abstract}

\section{Introduction}
\begin{figure}
\includegraphics[width=\linewidth]{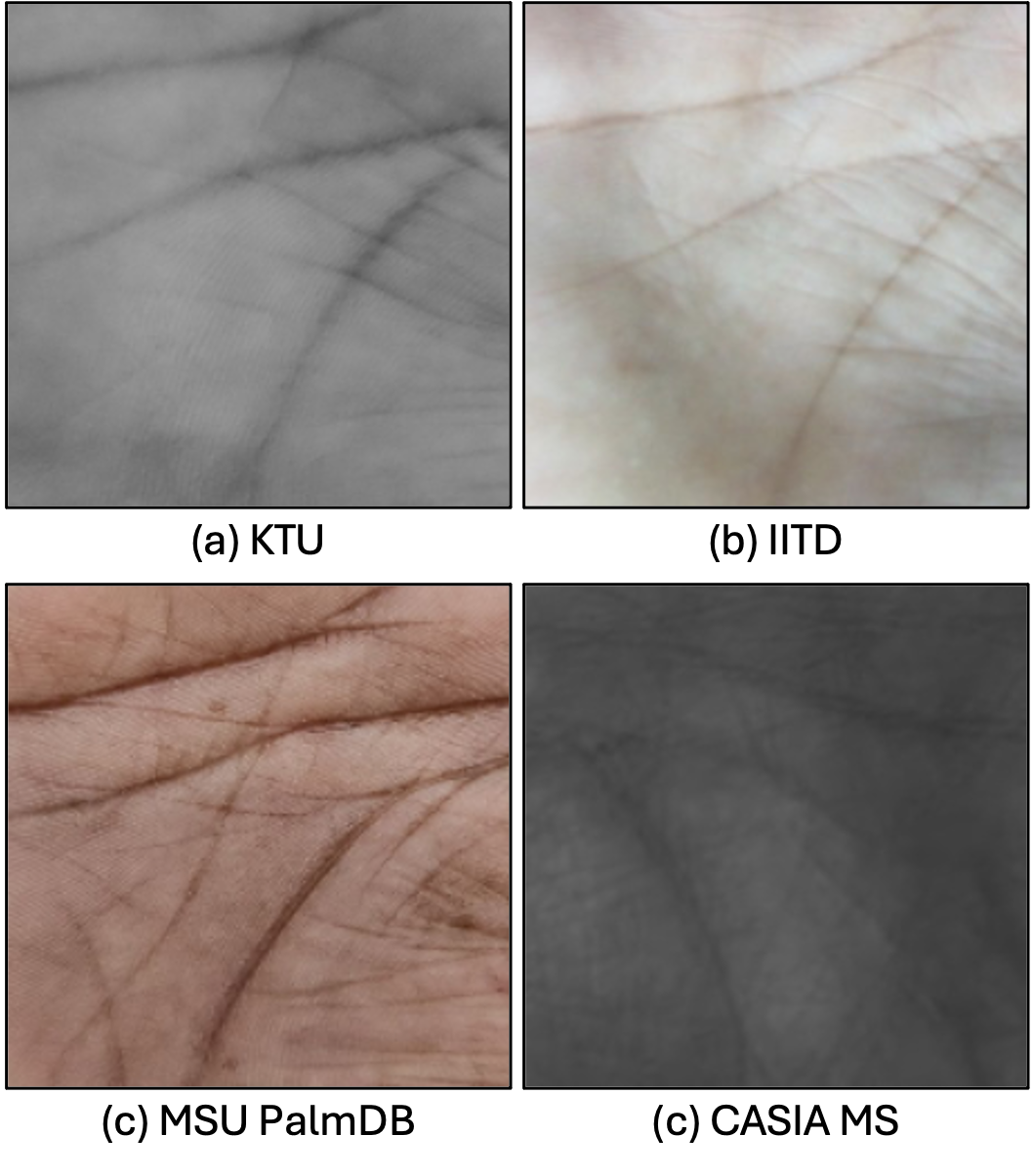} 
\caption{Example contactless palmprint region of interest (ROI) images from existing palmprint databases: (a) KTU~\cite{aykut2015developing}, (b) IITD v1~\cite{kumar2008incorporating, kumarpersonal}, (c) MSU PalmDB~\cite{grosz2024mobile}, and (d) CASIA Multispectral (CASIA MS)~\cite{hao2007comparative, hao2008multispectral}. ROIs are extracted at a resolution of 224x224 using the ROI extraction algorithm proposed in \cite{grosz2024mobile}.}
\label{fig:ex_real_imgs}
\end{figure}

Interest in contactless palmprint recognition has surged in recent years due to its high discriminability, ease of use, and lower privacy concerns compared to other biometric modalities. As a result, palmprint recognition deployments have grown in number, including those by major companies such as Amazon, Armatura, Fujitsu and Tencent~\cite{fei2018feature}. Deep learning-based recognition pipelines have become the mainstream technology in recent studies on contactless palmprint recognition~\cite{trabelsi2022efficient, godbole2023child, shao2024learning, shao2023privacy, zhu2023contactless, fei2022toward, yang2023co, yang2023comprehensive, yulin2023best, rong2022channel, grosz2024mobile}. However, a significant bottleneck in the research and application of deep learning-based palmprint recognition lies in the scarcity of large-scale palmprint datasets due to privacy concerns over the collection of palmprint images, such as those shown in Figure~\ref{fig:ex_real_imgs}.

To address this challenge, researchers have turned toward synthetic data generation to augment the limited amount of publicly available palmprint datasets. For example, BézierPalm~\cite{zhao2022bezierpalm} uses parameterized Bézier curves to simulate pseudo-palmprints and use them to pretrain recognition models. By adjusting the control points of Bézier curves according to geometric rules, BézierPalm can generate numerous new identities without requiring massive training data, contrasting with Generative Adversarial Network (GAN) approaches that do require extensive training datasets for good performance. Nonetheless, many GAN-based approaches for contactless palmprint generation have been proposed~\cite{minaee2020palm, chowdhury2023advancements, shen2023rpg, jin2024pce}. Specifically, Minaee \textit{et al.} build on a DC-GAN architecture, and Chowdhury \textit{et al.} employ a StyleGAN architecture for palmprint generation. Building on the previous work on Bézier curves, Shen \textit{et al.} and Jin \textit{et al.} further enhance the realism of Bézier-generated palmprints using GAN methods.

Despite the success of adversarial training mechanisms that enable GANs to generate highly realistic images, they suffer from two key problems. First, the training process can be unstable, requiring careful setting of hyperparameters. Second, there is the issue of mode collapse, where the generator produces only a limited range of output samples and fails to capture the full distribution of the training data.

Recently, diffusion probabilistic models~\cite{ho2020denoising, nichol2021improved} have emerged as state-of-the-art deep generative methods for image synthesis, effectively addressing the limitations of GANs. Research indicates that these models produce images of higher quality and greater stability~\cite{dhariwal2021diffusion}. Unlike GAN-based methods, diffusion probabilistic models progressively add fixed noise to training images, transforming them into Gaussian noise, and then focus solely on optimizing a denoising procedure to restore the degraded samples to their original form. This approach results in a stable training objective, leading to better distribution coverage and a reduced risk of mode collapse~\cite{ho2020denoising}.

In this paper, we propose a diffusion probabilistic model-based end-to-end contactless palmprint synthesis approach. Our main contributions are as follows:
 \begin{enumerate}
     \item To the best of our knowledge, this is the first study to introduce DDPMs for palmprint generation.
     \item We develop an end-to-end framework for contactless palmprint generation that synthesizes novel palm identities with multiple images per identity.
     \item We validate the realism and utility of the proposed synthetic palmprint images through several experiments, demonstrating their effectiveness in improving contactless palmprint recognition performance.
     \item Upon publication, we plan to release a large-scale synthetic palmprint database consisting of 12,000 unique palm identities with 20 images per palm.
 \end{enumerate}

\section{Related Work}
Synthetic palmprint generation can be broadly classified into two methodologies: (i) knowledge-driven and (ii) data-driven or learning-based methods. In this section, we briefly introduce both methods and provide a review of Denoising Diffusion Probabilistic Models (DDPMs).

\subsection{Previous Palmprint Generation Methods}
Synthetic contactless palmprint generation methods utilizing knowledge-driven processes began with Zhao \textit{et al.}~\cite{zhao2022bezierpalm}, who proposed synthesizing palmprints by manipulating palmar creases. They introduced an intuitive and simple geometric model to create palm images, representing palmar creases with parameterized Bézier curves. The identity of each synthesized sample is controlled by the parameters of these curves, such as the number of curves, positions of endpoints, and control points. This method can generate a vast number of samples with diverse identities, enabling large-scale pretraining on synthetic data. Models pretrained on this synthetic data demonstrated promising generalization ability and could be efficiently transferred to real datasets. However, the generated palm creases exhibited limited realism compared to actual palmprint images, prompting researchers to explore deep learning-based approaches for higher visual similarity to real images.

Shen \textit{et al.} built upon the ideas of Bézier palm crease generation by introducing a second stage in the generation process, translating the generated creases into realistic palmprint images using a GAN-based method. Taking it a step further, Jin \textit{et al.} introduced an intermediate stage to preprocess the Bézier curves with a higher degree of realism before applying realistic palmprint textures via GANs. This preprocessing stage connected the main principal lines and added smaller, realistic lines present in natural palmprint images. Besides extending the Bézier curve approach, other GAN-based methods have been proposed, such as extensions of DC-GAN by Minaee \textit{et al.} and StyleGAN by Jin \textit{et al.} for generating novel synthetic palmprint images. Despite the success achieved by these methods, the realism and utility of the generated palmprint images can be further improved by the more recent generation approach of Diffusion Probabilistic Models (DDPMs), which is the main focus of this paper.

\subsection{Denoising Diffusion Probabilistic Models}
Since the surge of interest in 2020, DDPMs~\cite{ho2020denoising, nichol2021improved} have achieved remarkable success across various domains, including natural language processing~\cite{yu2022latent}, image generation~\cite{rombach2022high}, and multi-modal modeling~\cite{avrahami2022blended}. Recently, DDPMs have also been applied in the biometrics community for generating synthetic face~\cite{kim2023dcface, huang2023collaborative}, fingerprint~\cite{li2023diffusion, grosz2024universal}, and iris~\cite{li2024i3fdm} images.

The DDPM generation process comprises two stages: a forward process that progressively perturbs data with additive noise towards a simple prior distribution, and a reverse process that gradually denoises the samples using a trainable deep neural network. During inference, a random noise vector is sampled from the prior distribution and transformed into a natural-looking image through the reverse denoising process. This progressive approach allows DDPMs to measure the likelihood directly and ensures much more stable training.

In this paper, we propose an end-to-end approach for contactless palmprint synthesis, designed to directly generate the palm crease structure with realistic intra-class variation, learned through the DDPM process.

\begin{figure*}
\includegraphics[width=\linewidth]{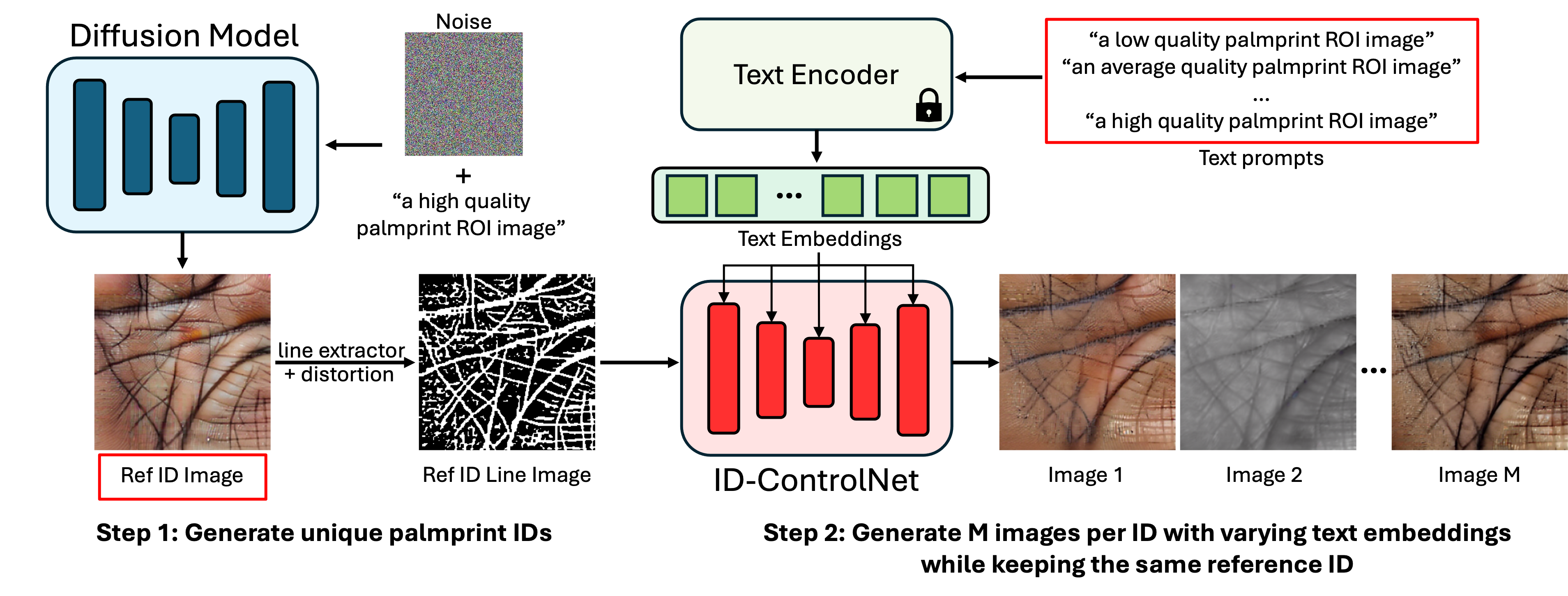} 
\caption{Overview of GenPalm architecture. GenPalm is able to replicate the diversity in color and textures present in its training corpus, demonstrated in the example outputs of the ID-ControlNet module.}
\label{fig:overview}
\end{figure*}

\section{Proposed Method}
An overview of the proposed DDPM-based palmprint generation method, GenPalm, is presented in Figure~\ref{fig:overview}. Our approach follows a two-stage process. In the first stage, diverse, novel palm identities are generated using a finetuned Stable Diffusion model~\cite{rombach2022high}. In the second stage, multiple images per palm are rendered by a finetuned ControlNet~\cite{zhang2023adding} model with identity preservation. Specific details regarding both stages are provided in the following subsections. This two-stage approach enables GenPalm to generate multiple images per palm identity with varying style characteristics.

\subsection{Stage One: Palmprint Identitiy Generation}
Step one of our generation pipeline consists of a Stable Diffusion model finetuned for contactless palmprint generation from the pretrained Stable Diffusion v1.5 weights made available from the open-source Diffusers library~\cite{von-platen-etal-2022-diffusers}. The forward process of the model is fixed, with no trainable parameters, and gradually adds Gaussian noise to the training palmprint images, transforming the data distribution into the standard Gaussian distribution. The reverse process, controlled by a learnable U-Net type architecture, converts noised samples from the standard Gaussian distribution into synthetic palmprint images through a step-by-step denoising process using the typical mean squared error objective employed by DDPMs. To synthesize palmprint images during inference, we feed in a text prompt of ``a high quality palmprint ROI image" along with a randomly sampled noise vector for each uniquely generated palm identity.

For training the stage one diffusion model, we constructed a dataset consisting of palmprint images and associated text prompt labels utilizing the following template: ``a \{quality\} palmprint ROI image", where the quality labels for each image were obtained from a pretrained Palm-ID model~\cite{grosz2024mobile}, which utilized the the L2 norm of the embeddings as a quality measure. The training set details are given in Table~\ref{tab:datasets}, which consisted of 7,873 unique palm identities and 165,927 total palmprint images. We extract the ROIs at a resolution of 512x512 from each palmprint image using the ROI extraction proposed in \cite{grosz2024mobile}. For fine-tuning Stable Diffusion on our text to palmprint image dataset, we utilize the low-rank adaptation (LoRA) strategy for more efficient training with a rank of 128~\cite{hu2022lora}. The LoRA weights are finetuned with a learning rate of 0.0001, cosine scheduler~\cite{loshchilov2016sgdr}, default Adam optimizer~\cite{kingma2014adam}, and batch size of 96 spread across 8 Nvidia A100 GPUs. The model is trained for 88,000 steps and trained on palmprint ROI images of a resolution of $512\times512$ pixels.

\begin{table*}
\centering
{\small
\begin{threeparttable}
    \caption{Summary of contactless palmprint databases used in this study. Training and test splits are disjoint.}
    \begin{tabular}{lcccc}
        \toprule
        \multicolumn{5}{c}{\textbf{Training Databases}} \\
        \midrule
        Name & \# Unique Palms & \# Images & Capture Device & Time-separated Collection \\
        \midrule
        MSU PalmDB~\cite{grosz2024mobile}\tnote{1} & 4,185 & 91,376 & Samsung Galaxy S22 & Yes (13 mos.) \\
        CASIA Multispectral~\cite{hao2007comparative, hao2008multispectral} & 200 & 7,200 & Custom Sensor & No \\
        Tongji~\cite{zhang2017towards} & 600 & 12,000 & Custom Sensor & Yes (2 mos.) \\
        11K Hands~\cite{afifi201911kHands} & 376 & 5,396 & n.a. & No \\
        Proprietary Database & 2,128 & 45,424 & n.a. & No \\
        SMPD~\cite{izadpanahkakhk2019novel} & 91 & 1,610 & DSLR Camera & No \\
        COEP~\cite{coep} & 146 & 1,169 & DSLR Camera & No \\
        KTU~\cite{aykut2015developing} & 147 & 1,752 & Flatbed Scanner & No \\
        \midrule
        \textbf{Total} & \textbf{7,873} & \textbf{165,927} & \textbf{-} & \textbf{-} \\
        \midrule
        \multicolumn{5}{c}{\textbf{Test Databases}} \\
        \midrule
        Name & \# unique palms & \# images & Capture Device & Time-Separated collection \\
        \midrule
        MSU CPDB-2-3~\cite{grosz2024mobile} & 258 & 10,031 & Samsung Galaxy S22 & Yes (7 mos.) \\
        MSU APDB-2-3~\cite{grosz2024mobile} & 406 & 15,424 & Samsung Galaxy S22 & Yes (7 mos.) \\
        CASIA Palmprint~\cite{sunordinal} & 620 & 5,502 & Custom Sensor& No \\
        IITD v1~\cite{kumar2008incorporating, kumarpersonal} & 460 & 2,601 & n.a. & No \\
        NTU Controlled~\cite{matkowski2019palmprint} & 655 & 2,478 & DSLR Camera & No \\
        \midrule
        \textbf{Total} & \textbf{2,399} & \textbf{36,036} & \textbf{-} & \textbf{-} \\
        \bottomrule
        \end{tabular}
    \begin{tablenotes}
    \item[1] Excluding MSU CPDB2-3 and MSU APDB-2-3.
    \end{tablenotes}
    \label{tab:datasets}
\end{threeparttable}
}
\end{table*}

\subsection{Stage Two: Identity Preserving ControlNet}
Several techniques have been proposed to maintain identity and personalize diffusion models. For instance, methods like Textual Inversion~\cite{gal2022image} and DreamBooth~\cite{ruiz2023dreambooth} require additional fine-tuning of networks to grasp new concepts tied to specific tokens. On the other hand, approaches like IP-Adapter~\cite{wang2024instantid} and PhotoMaker~\cite{li2023photomaker} utilize image embedding inputs to generate consistent identities across multiple subjects without the need for fine-tuning during inference. Both IP-Adapter and PhotoMaker integrate the identity of input reference images into the diffusion process through cross-attention layers, guiding the model to produce identity-consistent generations.

In contrast, methods like ControlNet aim to achieve precise spatial control over generation by directly introducing extra input in the image space. This strategy, successful in face~\cite{huang2023collaborative} and fingerprint~\cite{grosz2024universal} domains, provides detailed control during image synthesis. Given that palmprint identities are primarily defined by the structural features of prominent line creases due to palm folding and genetic characteristics, we propose leveraging a control image comprising palm crease lines for identity preservation in diffusion-based palmprint generation. The distinctiveness of palmprint identities lies in the consistent silhouettes of line patterns across various acquisition and sensor types. Our solution, ID-ControlNet, adapts the ControlNet framework by integrating a line extraction module into the identity preservation process. This module, based on image processing techniques, extracts the palm crease pattern from the input control image, removing sensor-specific and stylistic attributes. To extract the lines, the input palmprint images are first smoothed with a Gaussian blur with kernel size of 5, binarized using an adaptive thresholding algorithm, and post-processed with a morphological closing operation. An example input palmprint image and its extracted line image are shown in Figure~\ref{fig:overview}.

\begin{figure*}
\includegraphics[width=\linewidth]{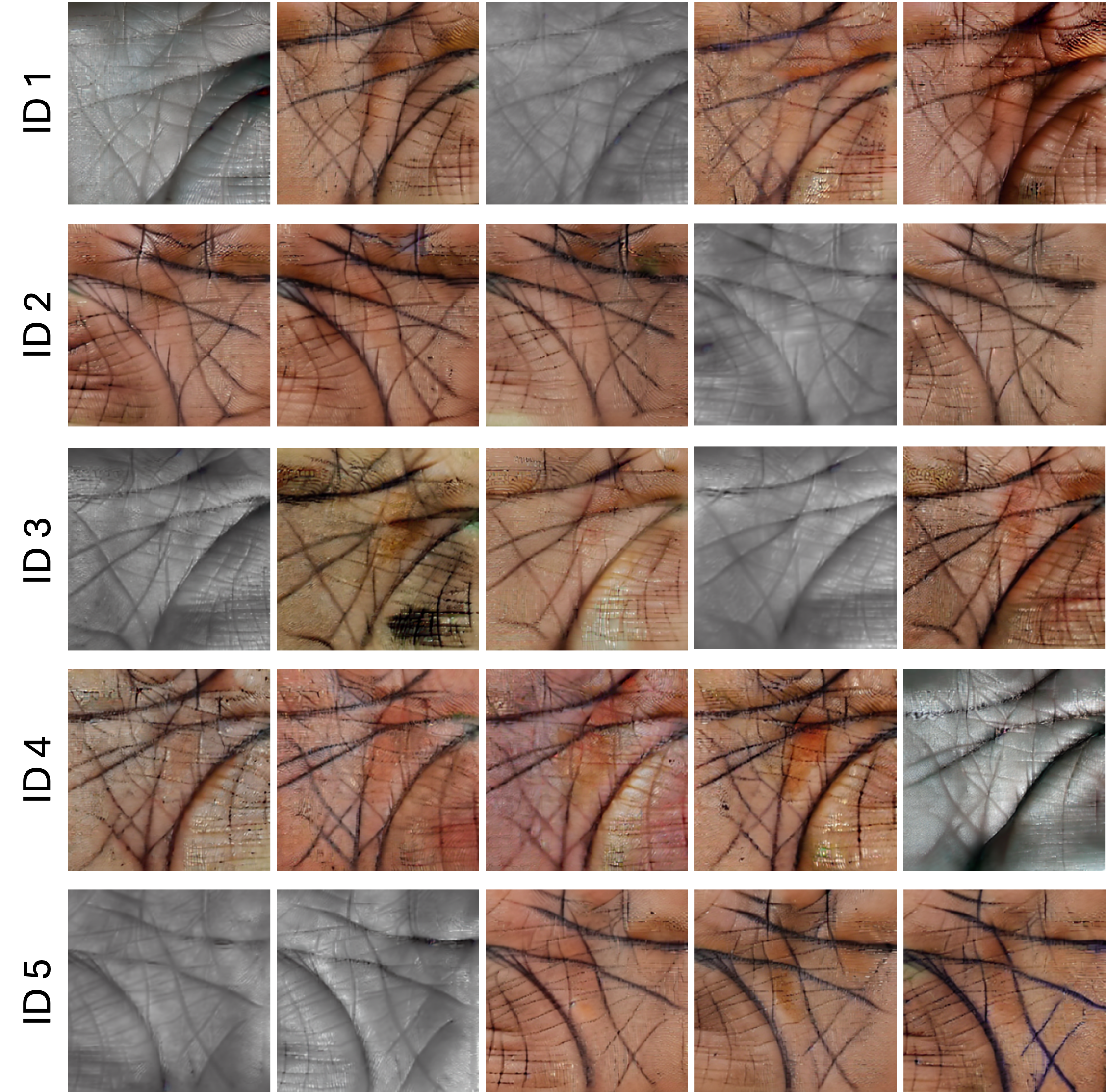} 
\caption{Example GenPalm-generated images of five different synthetic identities with five images per identity. GenPalm is able to generate highly realistic palmprint ROIs with useful intra-class and inter-class separation, including various color and texture variations.}
\label{fig:ex_imgs}
\end{figure*}

An additional input to the ID-ControlNet model is the text embedding prompt to further refine the generated images' quality, ensuring that ID-ControlNet produces varied textural characteristics while preserving the input palmprint crease pattern. Given the diversity in color and textures within our training database, ID-ControlNet is able to replicate those various intra-class variations in its generated outputs. The ID-ControlNet model weights are finetuned with a learning rate of 0.0001, cosine scheduler~\cite{loshchilov2016sgdr}, default Adam optimizer~\cite{kingma2014adam}, and batch size of 16 spread across 8 Nvidia A100 GPUs. The model is trained for 1,700 steps at a resolution of $512\times512$ pixels starting from the pretrained LoRA weights of the Stable Diffusion model from stage 1. Example rendered images of five different synthetically generated palmprint identities with five images per identity are shown in Figure~\ref{fig:ex_imgs} to demonstrate the realism and useful intra-class and inter-class separation of GenPalm generated images.

Lastly, to enhance the realism and utility of GenPalm-generated images, we incorporate perspective geometric distortions into the generation process. These distortions are derived from images in real palmprint datasets corresponding to the same palm. We achieve this by calculating perspective transformations between genuine pairs in our real training dataset. This involves extracting Oriented FAST and Rotated BRIEF (ORB) keypoints~\cite{rublee2011orb} and using the RANSAC~\cite{fischler1981random} algorithm to match corresponding keypoints. From these matched keypoints, we compile a dictionary of homographic transformation parameters, which are sampled randomly during GenPalm's palmprint generation process. These homographic transformations are then applied to the ControlNet input during the second stage of the generation process.

\section{Experiments}
In this section, we first assess the realism of GenPalm-generated images in comparison to actual contactless palmprint images. Subsequently, we gauge the effectiveness of synthetic images for training palmprint recognition models. This evaluation includes both training solely on synthetic images and augmenting real palmprint datasets with synthetic identities. 

\subsection{Realism of Synthetic Palms}
\begin{figure}
\includegraphics[width=\linewidth]{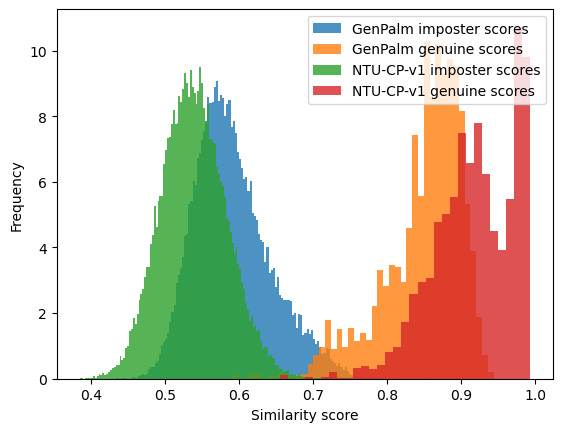} 
\caption{Pretrained Palm-ID~\cite{grosz2024mobile} similarity score distributions for the real NTU-CP-v1 palmprint database and corresponding sized GenPalm-generated database. Due to the larger intra-class variations present in GenPalm images, the imposter scores and genuine score distributions are slightly more overlapped than the real dataset.}
\label{fig:realism}
\end{figure}

To validate the realism of GenPalm-generated images we first compute similarity score distributions using a pertrained Palm-ID model~\cite{grosz2024mobile} to compare the score distributions between GenPalm-generated images and real palmprint images from the NTU-CP-v1 dataset. These score distributions are shown in Figure~\ref{fig:realism} which shows high similarity between the genuine and imposter distributions; however, the imposter and genuine score distributions for GenPalm are slightly more overlapped due to the larger intra-class variations present in GenPalm-generated images.

\begin{figure}
\includegraphics[width=\linewidth]{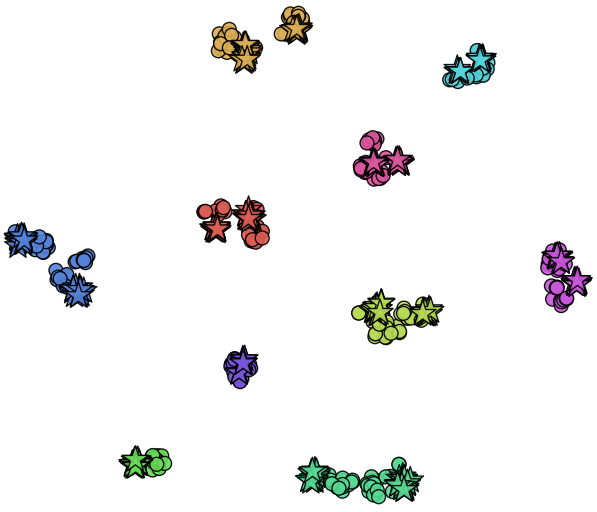} 
\caption{T-SNE embeddings of real palmprint images and GenPalm synthetic images of the same 10 identities (denoted by different colors). The proximity of real (denoted by dots) and synthetic (denoted by stars) embeddings corresponding to the same identity (i.e., color) validates both the realism and identity preservation of GenPalm-generated images.}
\label{fig:tsne}
\end{figure}

Next, to further show the realism of GenPalm-generated images, we take 10 example real palm identities from the test set of MSU PalmDB APDB-2-3 dataset and use the second stage of our GenPalm architecture to generate synthetic images of each identity. Then, we extract Palm-ID embeddings from approximately 20 real images of each ID and 20 synthetic images of each identity and plot them in the t-SNE ~\cite{van2008visualizing} embedding space to show the overlap between corresponding real and synthetic images. The closeness between synthetic and real images of the same identity in Figure~\ref{fig:tsne} further validates the realism of GenPalm-generated images.

\subsection{Utility for Recognition Model Training}
In this section, we present the experimental results of our proposed diffusion probabilistic model-based approach for contactless palmprint recognition. We compare the performance of models trained with different training data configurations, including real palmprint images and GenPalm-generated synthetic palmprint images.

We train both ResNet50~\cite{he2016deep} and ViT small~\cite{dosovitskiy2020image} architectures for a cross-model analysis. The real training dataset details are given in Table~\ref{tab:datasets} and consists of roughly 8,000 identities and 166,000 images (approximately 20 images per identity). For the experiments we randomly sample intervals of 1,000, 4,000, and 8,000 identities to analyze the performance as the number of real and synthetic training identities increases. For these experiments, we generated up to 12,000 unique synthetic identities with 20 images per identity using GenPalm. All recognition models are trained on two Nvidia RTX A6000 TI GPUs, with a total batch size of 512, Adam~\cite{kingma2014adam} optimizer, polynomial learning rate decay schedule, weight decay of $2e^{-5}$, and trained for a max of 200 epochs. An intial learning rate of 0.001 and 0.0001 were used for ResNet50 and ViT, respectively. 

\begin{figure*}
\includegraphics[width=\linewidth]{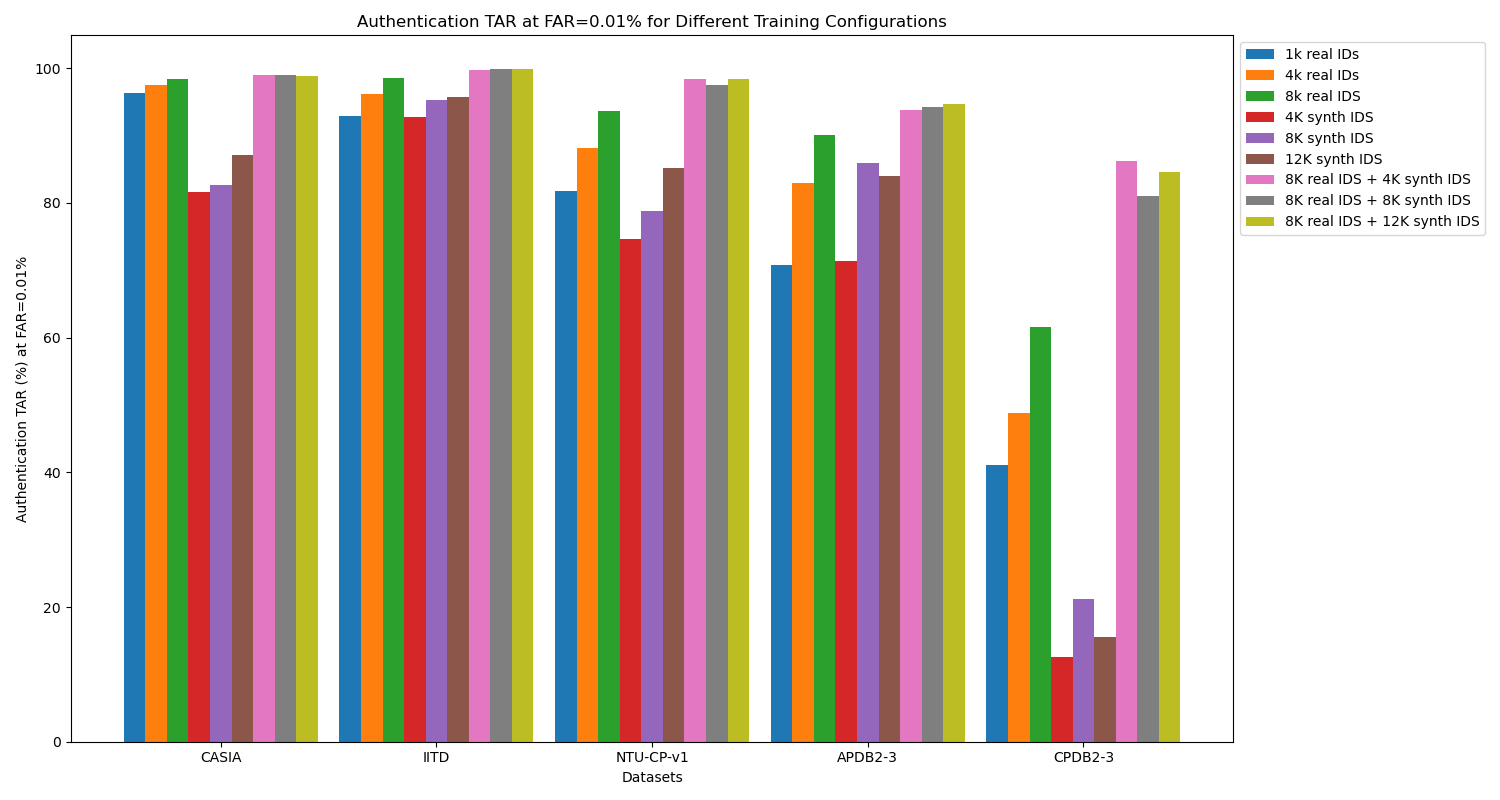} 
\caption{Performance comparison of palmprint recognition models with different training data configurations of real and GenPalm-generated synthetic palmprint images on five different test sets. Results reported in TAR (\%) at FAR=0.01\%.}
\label{fig:performance}
\end{figure*}

Table~\ref{tab:performance} presents the performance comparison of models with various training data configurations, reporting True Acceptance Rate (TAR) at a False Acceptance Rate (FAR) of 0.01\%. The models are evaluated on different palmprint databases, namely CASIA, NTU-CP-v1, APDB2-3, and CPDB2-3 (dataset statistics are given in Table~\ref{tab:datasets}). The results demonstrate the effectiveness of our proposed approach in synthesizing palmprint images and improving recognition performance. In some cases, such as IITD, NTU-CP-v1, and APDB-2-3, training on 12,000 GenPalm synthetic palmprint identities performs on par with training on 4,000 real palmprint identities. Overall, the best performance is achieved with a combination of 8,000 real and 12,000 synthetic palm identities. A visual representation of the performance comparison is provided in Figure~\ref{fig:performance}.

\begin{table*}
\centering
\caption{Performance comparison of palmprint recognition models with different training data configurations of real and GenPalm-generated synthetic palmprint images. Results reported in TAR (\%) at FAR=0.01\%.}
\renewcommand{\arraystretch}{1.3} 
\setlength{\tabcolsep}{8pt} 
\resizebox{\textwidth}{!}{%
\begin{tabular}{llcccccc}
\toprule
\textbf{Model} & \textbf{Train Data Description} & \textbf{CASIA} & \textbf{IITD} & \textbf{NTU-CP-v1} & \textbf{APDB2-3} & \textbf{CPDB2-3} & \textbf{Average} \\
\toprule
ResNet50 & 1k real IDs & 96.32 & 92.94 & 81.76 & 70.82 & 41.12 & 76.99 \\
ResNet50 & 4k real IDs & 97.46 & 96.22 & 88.12 & 83.03 & 48.81 & 82.73 \\
ResNet50 & 8k real IDs & 98.36 & 98.51 & 93.7 & 90.11 & 61.58 & 88.05 \\ 
\toprule
ResNet50 & 4K GenPalm IDs & 81.61 & 92.83 & 74.64 & 71.37 & 12.62 & 66.61 \\
ResNet50 & 8K GenPalm IDs & 82.7 & 95.23 & 78.73 & 85.89 & 21.17 & 72.74 \\
ResNet50 & 12K GenPalm IDs & 87.18 & 95.76 & 85.23 & 84.07 & 15.51 & 73.55 \\ \toprule
ResNet50 & 8K real IDs + 4K GenPalm IDs & 99.02 & 99.79 & 98.4 & 93.85 & \textbf{86.2} & 95.05 \\
ResNet50 & 8K real IDs + 8K GenPalm IDs & \textbf{99.03} & \textbf{99.89} & 97.45 & 94.18 & 80.99 & 94.71 \\
ResNet50 & 8K real IDs + 12K GenPalm IDs & 98.89 & 99.87 & \textbf{98.35} & \textbf{94.65} & 84.59 & \textbf{95.67} \\
\toprule
ViT Small & 8K real IDs & 96.51 & 96.07 & 90.25 & 87.11 & 63.29 & 86.25 \\
ViT Small & 12K GenPalm IDs & 90.99 & 96.36 & 85.77 & 88.35 & 40.1 & 80.71 \\
ViT Small & 8K real IDs + 12K GenPalm IDs & 98.36 & 98.44 & 95.38 & 90.88 & 72.72 & 91.16 \\ 
\bottomrule
\end{tabular}}
\label{tab:performance}
\end{table*}

\section{Conclusion}
In this study, we employed state-of-the-art diffusion model-based synthesis methods to develop GenPalm, a two-stage contactless palmprint generation approach. This method not only enhances the realism of the generated images but also increases their diversity, leveraging the ability of DDPMs to more effectively capture the full training distribution compared to previous GAN-based methods. To validate the realism of GenPalm-generated images, we compared the similarity score distributions of a pretrained palmprint recognition model with those of a real palmprint database. Additionally, we showcased the closeness between GenPalm-generated images and real images of corresponding palmprint identities in the t-SNE embedding space of a pretrained palmprint recognition model. Crucially, we demonstrated the utility of GenPalm synthetic images in improving the performance of palmprint recognition models. The findings of this study, along with the large-scale synthetic database we will release publicly, will significantly address the scarcity of publicly available contactless palmprint datasets and help advance the field of contactless palmprint recognition.

{\small
\bibliographystyle{ieee}
\bibliography{egbib}

\begin{thebibliography}{10}\itemsep=-1pt

\bibitem{afifi201911kHands}
M.~Afifi.
\newblock {11K Hands: Gender Recognition and Biometric Identification Using a Large Dataset of Hand Images}.
\newblock {\em Multimedia Tools and Applications}, 2019.

\bibitem{avrahami2022blended}
O.~Avrahami, D.~Lischinski, and O.~Fried.
\newblock Blended diffusion for text-driven editing of natural images.
\newblock In {\em Proceedings of the IEEE/CVF Conference on Computer Vision and Pattern Recognition}, pages 18208--18218, 2022.

\bibitem{aykut2015developing}
M.~Aykut and M.~Ekinci.
\newblock {Developing a Contactless Palmprint Authentication System by Introducing a Novel ROI Extraction Method}.
\newblock {\em Image and Vision Computing}, 40:65--74, 2015.

\bibitem{chowdhury2023advancements}
A.~M. Chowdhury, M.~J.~A. Khondkar, and M.~H. Imtiaz.
\newblock Advancements in synthetic generation for contactless palmprint biometrics using stylegan2-ada and stylegan3.
\newblock {\em Preprints}, 2023.

\bibitem{coep}
COEP.
\newblock {COEP} {P}alm {P}rint {D}atabase.
\newblock https://www.coep.org.in/resources/coeppalmprintdatabase.

\bibitem{dhariwal2021diffusion}
P.~Dhariwal and A.~Nichol.
\newblock Diffusion models beat gans on image synthesis.
\newblock {\em Advances in neural information processing systems}, 34:8780--8794, 2021.

\bibitem{dosovitskiy2020image}
A.~Dosovitskiy, L.~Beyer, A.~Kolesnikov, D.~Weissenborn, X.~Zhai, T.~Unterthiner, M.~Dehghani, M.~Minderer, G.~Heigold, S.~Gelly, et~al.
\newblock An image is worth 16x16 words: Transformers for image recognition at scale.
\newblock {\em arXiv preprint arXiv:2010.11929}, 2020.

\bibitem{fei2018feature}
L.~Fei, G.~Lu, W.~Jia, S.~Teng, and D.~Zhang.
\newblock Feature extraction methods for palmprint recognition: A survey and evaluation.
\newblock {\em IEEE Transactions on Systems, Man, and Cybernetics: Systems}, 49(2):346--363, 2018.

\bibitem{fei2022toward}
L.~Fei, S.~Zhao, W.~Jia, B.~Zhang, J.~Wen, and Y.~Xu.
\newblock {Toward Efficient Palmprint Feature Extraction by Learning a Single-Layer Convolution Network}.
\newblock {\em IEEE Transactions on Neural Networks and Learning Systems}, 2022.

\bibitem{fischler1981random}
M.~A. Fischler and R.~C. Bolles.
\newblock Random sample consensus: a paradigm for model fitting with applications to image analysis and automated cartography.
\newblock {\em Communications of the ACM}, 24(6):381--395, 1981.

\bibitem{gal2022image}
R.~Gal, Y.~Alaluf, Y.~Atzmon, O.~Patashnik, A.~H. Bermano, G.~Chechik, and D.~Cohen-Or.
\newblock An image is worth one word: Personalizing text-to-image generation using textual inversion.
\newblock {\em arXiv preprint arXiv:2208.01618}, 2022.

\bibitem{godbole2023child}
A.~Godbole, S.~A. Grosz, and A.~K. Jain.
\newblock {Child Palm-ID: Contactless Palmprint Recognition for Children}.
\newblock {\em arXiv preprint arXiv:2305.05161}, 2023.

\bibitem{grosz2024mobile}
S.~A. Grosz, A.~Godbole, and A.~K. Jain.
\newblock Mobile contactless palmprint recognition: Use of multiscale, multimodel embeddings.
\newblock {\em arXiv preprint arXiv:2401.08111}, 2024.

\bibitem{grosz2024universal}
S.~A. Grosz and A.~K. Jain.
\newblock Universal fingerprint generation: Controllable diffusion model with multimodal conditions.
\newblock {\em arXiv preprint arXiv:2404.13791}, 2024.

\bibitem{hao2007comparative}
Y.~Hao, Z.~Sun, and T.~Tan.
\newblock {Comparative Studies on Multispectral Palm Image Fusion for Biometrics}.
\newblock In {\em Asian conference on computer vision (ACCV)}, pages 12--21. Springer, 2007.

\bibitem{hao2008multispectral}
Y.~Hao, Z.~Sun, T.~Tan, and C.~Ren.
\newblock Multispectral {P}alm {I}mage {F}usion for {A}ccurate {C}ontact-{F}ree {P}almprint {R}ecognition.
\newblock In {\em 15th IEEE ICIP}, 2008.

\bibitem{he2016deep}
K.~He, X.~Zhang, S.~Ren, and J.~Sun.
\newblock Deep residual learning for image recognition.
\newblock In {\em Proceedings of the IEEE conference on computer vision and pattern recognition}, pages 770--778, 2016.

\bibitem{ho2020denoising}
J.~Ho, A.~Jain, and P.~Abbeel.
\newblock Denoising diffusion probabilistic models.
\newblock {\em Advances in neural information processing systems}, 33:6840--6851, 2020.

\bibitem{hu2022lora}
E.~J. Hu, Y.~Shen, P.~Wallis, Z.~Allen-Zhu, Y.~Li, S.~Wang, L.~Wang, and W.~Chen.
\newblock Lo{RA}: Low-rank adaptation of large language models.
\newblock In {\em International Conference on Learning Representations}, 2022.

\bibitem{huang2023collaborative}
Z.~Huang, K.~C. Chan, Y.~Jiang, and Z.~Liu.
\newblock Collaborative diffusion for multi-modal face generation and editing.
\newblock In {\em Proceedings of the IEEE/CVF Conference on Computer Vision and Pattern Recognition}, pages 6080--6090, 2023.

\bibitem{izadpanahkakhk2019novel}
M.~Izadpanahkakhk et~al.
\newblock {Novel Mobile Palmprint Databases for Biometric Authentication}.
\newblock {\em International Journal of Grid and Utility Computing}, 10(5):465--474, 2019.

\bibitem{jin2024pce}
J.~Jin, L.~Shen, R.~Zhang, C.~Zhao, G.~Jin, J.~Zhang, S.~Ding, Y.~Zhao, and W.~Jia.
\newblock Pce-palm: Palm crease energy based two-stage realistic pseudo-palmprint generation.
\newblock In {\em Proceedings of the AAAI Conference on Artificial Intelligence}, volume~38, pages 2616--2624, 2024.

\bibitem{kim2023dcface}
M.~Kim, F.~Liu, A.~Jain, and X.~Liu.
\newblock Dcface: Synthetic face generation with dual condition diffusion model.
\newblock In {\em Proceedings of the IEEE/CVF Conference on Computer Vision and Pattern Recognition}, pages 12715--12725, 2023.

\bibitem{kingma2014adam}
D.~P. Kingma and J.~Ba.
\newblock Adam: A method for stochastic optimization.
\newblock {\em arXiv preprint arXiv:1412.6980}, 2014.

\bibitem{kumar2008incorporating}
A.~Kumar.
\newblock {Incorporating Cohort Information for Reliable Palmprint Authentication}.
\newblock In {\em Sixth ICVGIP}. IEEE, 2008.

\bibitem{kumarpersonal}
A.~Kumar and S.~Shekhar.
\newblock {Personal Identification Using Rank-level Fusion}.
\newblock {\em IEEE Trans. Systems, Man, and Cybernetics: Part C}, pages 743--752, 2011.

\bibitem{li2024i3fdm}
C.~Li, Z.~Zhang, P.~Li, and Z.~He.
\newblock I3fdm: Iris inpainting via inverse fusion of diffusion models.
\newblock In {\em ICASSP 2024-2024 IEEE International Conference on Acoustics, Speech and Signal Processing (ICASSP)}, pages 1636--1640. IEEE, 2024.

\bibitem{li2023diffusion}
K.~Li and X.~Yang.
\newblock Diffusion probabilistic model based end-to-end latent fingerprint synthesis.
\newblock In {\em 2023 IEEE 4th International Conference on Pattern Recognition and Machine Learning (PRML)}, pages 343--349. IEEE, 2023.

\bibitem{li2023photomaker}
Z.~Li, M.~Cao, X.~Wang, Z.~Qi, M.-M. Cheng, and Y.~Shan.
\newblock Photomaker: Customizing realistic human photos via stacked id embedding.
\newblock {\em arXiv preprint arXiv:2312.04461}, 2023.

\bibitem{loshchilov2016sgdr}
I.~Loshchilov and F.~Hutter.
\newblock Sgdr: Stochastic gradient descent with warm restarts.
\newblock {\em arXiv preprint arXiv:1608.03983}, 2016.

\bibitem{matkowski2019palmprint}
W.~M. Matkowski et~al.
\newblock Palmprint {R}ecognition in {U}ncontrolled and {U}ncooperative {E}nvironment.
\newblock {\em IEEE Trans. IFS}, 2019.

\bibitem{minaee2020palm}
S.~Minaee, M.~Minaei, and A.~Abdolrashidi.
\newblock Palm-gan: Generating realistic palmprint images using total-variation regularized gan.
\newblock {\em arXiv preprint arXiv:2003.10834}, 2020.

\bibitem{nichol2021improved}
A.~Q. Nichol and P.~Dhariwal.
\newblock Improved denoising diffusion probabilistic models.
\newblock In {\em International conference on machine learning}, pages 8162--8171. PMLR, 2021.

\bibitem{rombach2022high}
R.~Rombach, A.~Blattmann, D.~Lorenz, P.~Esser, and B.~Ommer.
\newblock High-resolution image synthesis with latent diffusion models.
\newblock In {\em Proceedings of the IEEE/CVF conference on computer vision and pattern recognition}, pages 10684--10695, 2022.

\bibitem{rong2022channel}
W.~Rong, Z.~Yang, and L.~Leng.
\newblock Channel group-wise drop network with global and fine-grained-aware representation learning for palm recognition.
\newblock In {\em 2022 IEEE International Joint Conference on Biometrics (IJCB)}, pages 1--9. IEEE, 2022.

\bibitem{rublee2011orb}
E.~Rublee, V.~Rabaud, K.~Konolige, and G.~Bradski.
\newblock Orb: An efficient alternative to sift or surf.
\newblock In {\em 2011 International conference on computer vision}, pages 2564--2571. Ieee, 2011.

\bibitem{ruiz2023dreambooth}
N.~Ruiz, Y.~Li, V.~Jampani, Y.~Pritch, M.~Rubinstein, and K.~Aberman.
\newblock Dreambooth: Fine tuning text-to-image diffusion models for subject-driven generation.
\newblock In {\em Proceedings of the IEEE/CVF Conference on Computer Vision and Pattern Recognition}, pages 22500--22510, 2023.

\bibitem{shao2023privacy}
H.~Shao, C.~Liu, X.~Li, and D.~Zhong.
\newblock {Privacy Preserving Palmprint Recognition via Federated Metric Learning}.
\newblock {\em IEEE Transactions on Information Forensics and Security}, 2023.

\bibitem{shao2024learning}
H.~Shao, Y.~Zou, C.~Liu, Q.~Guo, and D.~Zhong.
\newblock {Learning to Generalize Unseen Dataset for Cross-dataset Palmprint Recognition}.
\newblock {\em IEEE Transactions on Information Forensics and Security}, 2024.

\bibitem{shen2023rpg}
L.~Shen, J.~Jin, R.~Zhang, H.~Li, K.~Zhao, Y.~Zhang, J.~Zhang, S.~Ding, Y.~Zhao, and W.~Jia.
\newblock Rpg-palm: Realistic pseudo-data generation for palmprint recognition.
\newblock In {\em Proceedings of the IEEE/CVF International Conference on Computer Vision}, pages 19605--19616, 2023.

\bibitem{sunordinal}
Z.~Sun, T.~Tan, Y.~Wang, and S.~Li.
\newblock Ordinal {P}almprint {R}epresentation for {P}ersonal {I}dentification.
\newblock In {\em Proceedings of the IEEE CVPR}, 2005.

\bibitem{trabelsi2022efficient}
S.~Trabelsi, D.~Samai, F.~Dornaika, A.~Benlamoudi, K.~Bensid, and A.~Taleb-Ahmed.
\newblock {Efficient Palmprint Biometric Identification Systems Using Deep Learning and Feature Selection Methods}.
\newblock {\em Neural Computing and Applications}, 34(14):12119--12141, 2022.

\bibitem{van2008visualizing}
L.~Van~der Maaten and G.~Hinton.
\newblock Visualizing data using t-sne.
\newblock {\em Journal of machine learning research}, 9(11), 2008.

\bibitem{von-platen-etal-2022-diffusers}
P.~von Platen, S.~Patil, A.~Lozhkov, P.~Cuenca, N.~Lambert, K.~Rasul, M.~Davaadorj, D.~Nair, S.~Paul, W.~Berman, Y.~Xu, S.~Liu, and T.~Wolf.
\newblock Diffusers: State-of-the-art diffusion models.
\newblock \url{https://github.com/huggingface/diffusers}, 2022.

\bibitem{wang2024instantid}
Q.~Wang, X.~Bai, H.~Wang, Z.~Qin, and A.~Chen.
\newblock Instantid: Zero-shot identity-preserving generation in seconds.
\newblock {\em arXiv preprint arXiv:2401.07519}, 2024.

\bibitem{yang2023comprehensive}
Z.~Yang, H.~Huangfu, L.~Leng, B.~Zhang, A.~B.~J. Teoh, and Y.~Zhang.
\newblock Comprehensive competition mechanism in palmprint recognition.
\newblock {\em IEEE Transactions on Information Forensics and Security}, 2023.

\bibitem{yang2023co}
Z.~Yang, W.~Xia, Y.~Qiao, Z.~Lu, B.~Zhang, L.~Leng, and Y.~Zhang.
\newblock Co 3 net: Coordinate-aware contrastive competitive neural network for palmprint recognition.
\newblock {\em IEEE Transactions on Instrumentation and Measurement}, 2023.

\bibitem{yu2022latent}
P.~Yu, S.~Xie, X.~Ma, B.~Jia, B.~Pang, R.~Gao, Y.~Zhu, S.-C. Zhu, and Y.~N. Wu.
\newblock Latent diffusion energy-based model for interpretable text modeling.
\newblock {\em arXiv preprint arXiv:2206.05895}, 2022.

\bibitem{yulin2023best}
F.~Yulin and A.~Kumar.
\newblock {BEST: Building Evidences From Scattered Templates for Accurate Contactless Palmprint Recognition}.
\newblock {\em Pattern Recognition}, 138:109422, 2023.

\bibitem{zhang2017towards}
L.~Zhang, L.~Li, A.~Yang, Y.~Shen, and M.~Yang.
\newblock Towards {C}ontactless {P}almprint {R}ecognition: {A} {N}ovel {D}evice, a {N}ew {B}enchmark, and a {C}ollaborative {R}epresentation {B}ased {I}dentification {A}pproach.
\newblock {\em Pattern Recognition}, 69:199--212, 2017.

\bibitem{zhang2023adding}
L.~Zhang, A.~Rao, and M.~Agrawala.
\newblock Adding conditional control to text-to-image diffusion models.
\newblock In {\em Proceedings of the IEEE/CVF International Conference on Computer Vision}, pages 3836--3847, 2023.

\bibitem{zhao2022bezierpalm}
K.~Zhao, L.~Shen, Y.~Zhang, C.~Zhou, T.~Wang, R.~Zhang, S.~Ding, W.~Jia, and W.~Shen.
\newblock B{\'e}zierpalm: A free lunch for palmprint recognition.
\newblock In {\em European Conference on Computer Vision}, pages 19--36. Springer, 2022.

\bibitem{zhu2023contactless}
Q.~Zhu, G.~Xin, L.~Fei, D.~Liang, Z.~Zhang, D.~Zhang, and D.~Zhang.
\newblock {Contactless Palmprint Image Recognition across Smartphones with Self-paced CycleGAN}.
\newblock {\em IEEE Transactions on Information Forensics and Security}, 2023.

\end{thebibliography}
}

\end{document}